\documentclass[conference]{IEEEtran}
\IEEEoverridecommandlockouts

\usepackage{cite}
\usepackage{amsmath,amssymb,amsfonts}
\usepackage{algorithmic}
\usepackage{graphicx}
\usepackage{textcomp}
\usepackage{xcolor}
\usepackage{multirow}

\usepackage{pifont}
\usepackage{url}
\usepackage[hidelinks]{hyperref}

\usepackage{color, colortbl}
\definecolor{lightgreen}{rgb}{0.95,1.0,0.95}
\newcolumntype{a}{>{\columncolor{lightgreen}}c}

\def\BibTeX{{\rm B\kern-.05em{\sc i\kern-.025em b}\kern-.08em
    T\kern-.1667em\lower.7ex\hbox{E}\kern-.125emX}}

\title{WaspMOT: A Benchmark for Long-Term Multi-Object Tracking of \textit{Trichogramma} Wasps}
\author{
\IEEEauthorblockN{
Tomasz Stanczyk$^{1,3,*}$ \qquad
Yuan Gao$^{2,3,*}$
}

\vspace{0.2cm}

\IEEEauthorblockN{
Hardik Agarwal$^{1,4}$ \qquad
Seongro Yoon$^{1,3}$ \qquad
Tiantao Zhang$^{5}$ \qquad
Vincent Calcagno$^{2}$ \qquad
Francois Bremond$^{1}$
}

\vspace{0.2cm}

\IEEEauthorblockA{
$^{1}$Inria, Sophia Antipolis, France\\
$^{2}$INRAE Institut Sophia Agrobiotech, Sophia Antipolis, France\\
$^{3}$Université Côte d'Azur, Sophia Antipolis, France\\
$^{4}$Indian Institute of Technology Delhi, Delhi, India\\
$^{5}$Institute of Plant Protection, Chinese Academy of Agricultural Sciences, Beijing, China\\[0.3em]
$^{*}$Joint first authors.
}
}

\begin{document}
\maketitle

\begin{abstract}
Multi-object tracking (MOT) has achieved strong performance on benchmarks dominated by short video sequences. However, such datasets do not adequately evaluate long-term identity preservation, where objects must be tracked consistently over extended durations. We introduce \textbf{WaspMOT}, a benchmark designed to address this gap through long-duration tracking of \textit{Trichogramma} wasps in controlled ecological experiments. 
The dataset contains 10 sequences of approximately 12,000 frames each (over 8 minutes at 25 FPS), with dense MOTChallenge annotations and oracle detections to isolate association performance.

Unlike existing benchmarks, WaspMOT forms a closed-set tracking scenario where all individuals remain present throughout the sequence, requiring consistent identity assignment across thousands of frames despite abrupt jumps, occlusions, and highly similar appearance. We establish a benchmark by evaluating five tracking-by-detection methods, including ByteTrack, BoT-SORT, C-BIoU, OC-SORT, and McByte, under a unified protocol. Results show that all methods suffer from significant trajectory fragmentation, highlighting the difficulty of long-term identity preservation even with perfect detections. A simple spatial tracklet stitching baseline consistently improves performance, indicating that substantial gains remain possible.

WaspMOT provides a new benchmark for studying long-term association and reveals limitations of current tracking approaches that are not observable on conventional datasets. The benchmark will be made publicly available at the project repository: \href{https://github.com/tstanczyk95/WaspMOT/}{\texttt{https://github.com/tstanczyk95/WaspMOT/}}.
\end{abstract}


\section{Introduction}

Multi-object tracking (MOT) is a fundamental computer vision problem with applications in video surveillance, autonomous driving, robotics, and sports analytics. Its objective is to detect objects and maintain consistent identities over time. Recent progress has been driven by benchmark datasets focusing on human-centered scenarios, including pedestrian tracking~\cite{mot17_ref,mot20_ref}, sports tracking~\cite{sportsmot_ref,soccernet-tracking2022_ref}, and human motion analysis~\cite{dancetrack_ref}. These datasets have enabled robust tracking algorithms capable of handling occlusions, appearance changes, and moderate motion.

However, existing MOT benchmarks are dominated by short sequences, typically lasting tens of seconds, where trajectories span only a limited portion of the video. As a result, trackers are primarily evaluated on short-term association rather than long-term identity preservation. This limits the ability to study scenarios where identities must be maintained consistently over extended durations. In ecological and biological monitoring, such long-term identity consistency is essential for analyzing individual behaviors and interactions.

To address this gap, we introduce \textbf{WaspMOT}, a dataset designed for long-duration multi-object tracking of \textit{Trichogramma} wasps in laboratory-controlled experiments. Each sequence contains approximately 12,000 frames (over 8 minutes at 25 FPS), and all individuals remain present throughout the entire video, forming a closed-set tracking scenario. This setting enables evaluation of identity preservation across full-length trajectories spanning thousands of frames, a regime not covered by existing MOT benchmarks.

\begin{figure}[!t]
    \centerline{\includegraphics[width=1\linewidth]{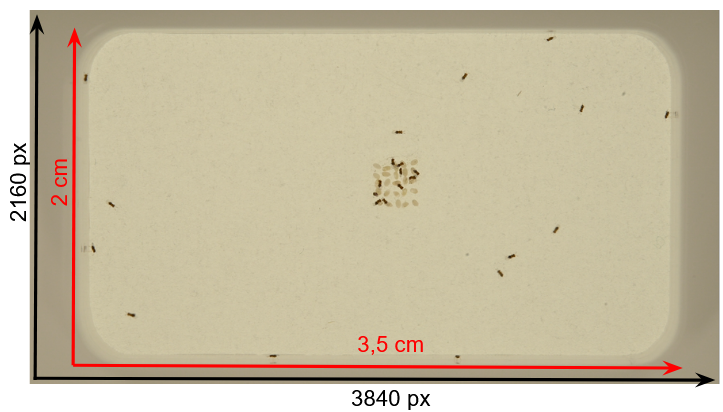}}
    \caption{Sample frame of a video with individuals inside an experimental arena. The image pixel dimensions are indicated in black and the real arena dimensions are indicated in red.}
    \label{fig:arena_dimensions}
\end{figure}

Tracking \textit{Trichogramma} wasps presents unique challenges. Individuals frequently perform abrupt jump events, causing large displacements and temporary disappearance from detections. Occlusions occur due to interactions and the three-dimensional arena structure, where wasps move on both floor and ceiling surfaces. In addition, individuals exhibit highly similar visual appearance, making appearance-based association difficult. These factors lead to frequent trajectory fragmentation even with perfect detections.

To focus on identity association, WaspMOT provides oracle detections derived from ground truth annotations in MOTChallenge format~\cite{mot17_ref}. This isolates tracking performance from detection errors and enables a controlled evaluation of long-term association. We establish a benchmark by evaluating five tracking-by-detection methods: ByteTrack~\cite{bt_ref}, BoT-SORT~\cite{botsort_ref}, C-BIoU~\cite{cbiou_ref}, OC-SORT~\cite{ocsort_ref}, and McByte~\cite{MCBYTE_REF}. 
These methods are evaluated under a unified protocol using the provided detections, enabling consistent comparison of their association capabilities.

Our results show that all evaluated trackers suffer from significant trajectory fragmentation under this setting, highlighting the difficulty of long-term identity preservation. To further analyze recoverable errors, we include a simple spatial tracklet stitching baseline that reconnects fragmented trajectories based on position and temporal consistency. This baseline consistently improves performance, indicating that substantial gains remain possible and motivating the development of more advanced long-term tracking approaches.

The contributions of this paper are as follows:

\begin{itemize}
\item We introduce WaspMOT, a benchmark dataset for long-term identity preservation with full-length trajectories in a closed-set tracking scenario.
\item We provide an evaluation protocol using oracle detections to isolate identity association performance.
\item We benchmark five tracking algorithms, showing that long-term identity preservation remains challenging even with perfect detections.
\item We include a simple spatial stitching baseline demonstrating recoverable fragmentation and motivating future research.
\item We will publicly release the dataset and code to support future work in long-term tracking and ecological surveillance.
\end{itemize}

\section{Related Work}

\subsection{Multi-Object Tracking Datasets}

MOT research has been driven primarily by pedestrian and human-centered benchmarks. The MOTChallenge datasets, including MOT17 and MOT20~\cite{mot17_ref,mot20_ref}, provide crowded urban scenarios with occlusions and dense interactions, but sequences typically span only short durations. Domain-specific datasets such as DanceTrack~\cite{dancetrack_ref}, SportsMOT~\cite{sportsmot_ref}, and SoccerNet-Tracking~\cite{soccernet-tracking2022_ref} extend evaluation to human motion and sports scenarios, introducing challenges such as rapid motion, interactions, and camera dynamics.

Despite their impact, existing MOT datasets consist mainly of short sequences lasting tens of seconds. As shown in Table~\ref{tab:dataset_statistics}, their average duration ranges from approximately 25 to 100 seconds, whereas WaspMOT sequences exceed 480 seconds. This substantial increase in temporal duration enables evaluation of long-term identity preservation, which is difficult to assess using conventional datasets.

Tracking has also been explored in ecological contexts. The BEE24 dataset~\cite{bee24_ref} provides an insect tracking benchmark, but trajectories typically cover only limited portions of each sequence. In contrast, WaspMOT provides a closed-set tracking scenario where all individuals remain present throughout the video, enabling evaluation of full-length trajectories.

\subsection{Tracking-by-Detection Methods}

Tracking-by-detection is the dominant MOT paradigm, where detections are associated across frames to maintain identities. Most approaches rely on motion prediction using Kalman filtering~\cite{kf_ref} combined with spatial association based on bounding box overlap.

Recent methods differ primarily in how they enhance this basic formulation. IoU-based approaches such as ByteTrack~\cite{bt_ref} emphasize robust association using detection confidence and spatial overlap. Extensions such as BoT-SORT~\cite{botsort_ref} incorporate appearance features to complement motion cues, while OC-SORT~\cite{ocsort_ref} focuses on improved motion modeling through observation-centric updates. Other methods, including C-BIoU~\cite{cbiou_ref}, refine spatial matching strategies to better handle fast motion. McByte~\cite{MCBYTE_REF} further extends this paradigm by incorporating temporally propagated segmentation masks~\cite{cutie_ref,sam_ref} to improve geometric consistency and resolve ambiguous associations.

When detections are provided externally, such methods can operate without additional detector training, enabling evaluation of pure association performance. However, despite these improvements, most approaches rely on motion continuity assumptions and struggle with abrupt motion or temporary disappearance, resulting in fragmented trajectories. Long-duration datasets such as WaspMOT expose these limitations more clearly than conventional benchmarks.

\subsection{Appearance-Based Association}

Appearance-based methods aim to improve identity association by learning discriminative representations of objects across time and viewpoints~\cite{market1501_ref,Wang_2022,Kongsilp_2024,Cermak_2024}. Deep models typically use convolutional neural networks trained with identity classification and metric learning objectives, such as triplet loss~\cite{triplet_loss_ref}, to produce feature embeddings~\cite{he2020fastreid,osnet_ref}. These embeddings are widely used in modern tracking pipelines to complement motion-based association~\cite{deepsort_ref,strongsort_ref,botsort_ref}.

However, their effectiveness depends on the availability of distinctive visual features. In scenarios such as insect tracking, where individuals are small and visually similar, appearance-based cues are inherently limited. This makes identity association particularly challenging and highlights the need for benchmarks that emphasize long-term consistency under such conditions.

\section{WaspMOT Dataset}

The WaspMOT dataset consists of laboratory-recorded videos of \textit{Trichogramma} wasps, annotated in the standard MOTChallenge format, and provides oracle detections to isolate tracking performance. WaspMOT is designed as a benchmark for long-term identity preservation, where sequence duration and full-length trajectories are the primary focus, rather than large-scale dataset size or detector training.

\subsection{Experimental Setup and Data Acquisition}

Experiments were performed in laboratory conditions, where \textit{Trichogramma} wasps are maintained in climate chambers and recorded inside an enclosed experimental arena of size 3.5cm×2cm, containing host eggs, as shown in Fig.~\ref{fig:arena_dimensions}. We consider two species and three strains: \textit{Ostriniae} (industrial strain) and \textit{Brassicae} (industrial and lab-reared strains), with differing behavior such as movements and jumps. Female individuals are of primary interest due to their role in host egg parasitism~\cite{Smith_1996,Wang_2024}. For each experiment, host eggs were manually placed at the center of the arena in a 5×5 grid to enable controlled observation of individual movements and interactions~\cite{Harvey_2013,Robert_2016}.

Recordings were performed at 25 FPS with a fixed overhead camera and last approximately 8 minutes each, resulting in at least 12,000 frames per video. The camera focus was set to the arena floor, producing a mixture of sharp and blurry observations depending on the wasps' position in the three-dimensional arena. The resolution of each video is 3840×2160 pixels.

\begin{table*}[!t]
\caption{Comparison of WaspMOT with existing multi-object tracking datasets. Unlike conventional benchmarks composed of short sequences, WaspMOT (in green) contains substantially longer videos with dense annotations and trajectories spanning the entire sequence duration. This enables evaluation of long-term identity preservation, which cannot be reliably assessed using existing datasets. '-' means that this information has not been specified in the source.}
\centering
\small
\begin{tabular}{|l|cccccca|}
\hline
Dataset & MOT17 & MOT20 & DanceTrack & SportsMOT & SoccerNet & BEE24 & WaspMOT (ours) \\
\hline
\# Videos                       & 14 & 8 & 100 & \textbf{240} & 201 & 36 & 10 \\
FPS                             & \textbf{27} (avg.) & 25 & 20 & 25 & 25 & 25 & 25 \\
\hline
Avg frames / video           & 802.50 & 1676.25 & 1,058.55 & 626.58 & 1121.26 & 654.17 & \textbf{12,033.00} \\
Total frames                    & 11,235 & 13,410 & 105,855 & 150,379 & \textbf{225,375} & 23,550 & 120,330 \\
\hline
Avg seconds / video          & 99.21 & 66.88 & 52.92 & 25.06 & 44.85 & 26.17 & \textbf{481.30} \\
Total seconds                   & 1,389 & 535 & 5,292 & 6,015 & \textbf{9,015} & 942 & 4,813 \\
\hline
Avg boxes / video            & 20,909.50 & 206,505 & - & 6789.54 & 18,137.62 & - & \textbf{256,952.70} \\
Total boxes                     & 292,733 & 1,652,040 & - & 1,629,490 & \textbf{3,645,661} & - & 2,569,527 \\
\hline
Avg tracks / video           & 95.86 & 432 & 9.9 & 14.17 & 24.92 & \textbf{126.64} & 21.40 \\
Total tracks                    & 1,342 & 3,456 & 990 & 3,401 & \textbf{5,009} & 4,559 & 214 \\
\hline
Full-video-long tracks & \ding{53} & \ding{53} & \ding{53} & \ding{53} & \ding{53} & \ding{53} & \ding{51} \\
\hline
\end{tabular}
\label{tab:dataset_statistics}
\end{table*}

\subsection{Dataset Composition and Annotation}

WaspMOT contains 10 fully annotated video sequences, with 5 sequences per species. Ground truth annotations are provided as bounding boxes in the standard MOTChallenge format~\cite{mot17_ref}, with a total of 120,330 annotated frames and 2,569,527 annotated object instances. Each sequence contains an average of 21.4 (varying from 15 to 28) individuals that remain present throughout the entire video duration, forming a closed-set tracking scenario with full-length trajectories.

Unlike conventional tracking datasets~\cite{mot17_ref,mot20_ref,dancetrack_ref,sportsmot_ref}, WaspMOT provides oracle detections derived directly from ground truth bounding boxes without identity labels. This enables focused evaluation of tracking performance independent of detection errors, as in~\cite{soccernet-tracking2022_ref}. Evaluation is performed using the TrackEval~\cite{trackeval_ref} toolkit, ensuring compatibility with standard MOT evaluation protocols.

A statistical comparison between WaspMOT and existing MOT tracking datasets is presented in Table~\ref{tab:dataset_statistics}. WaspMOT sequences are an order of magnitude longer than those in existing datasets, with an average duration exceeding 480 seconds per video. This enables evaluation of identity preservation across thousands of frames, a regime not captured by conventional benchmarks.

\begin{figure}[!t]
    \centering
    \centerline{\includegraphics[width=1\linewidth]{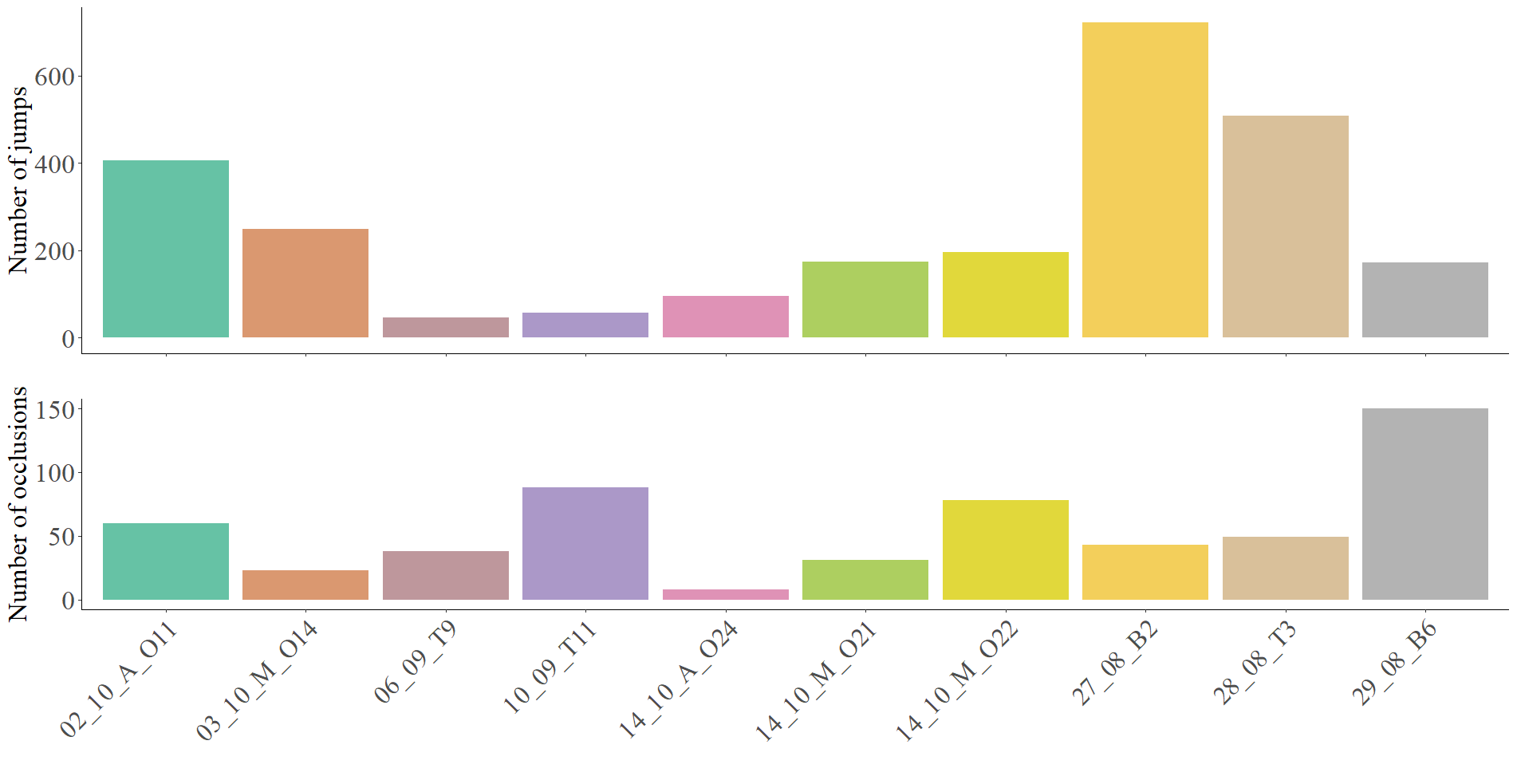}}
    \caption{Number of occlusion and jump events for each video. Original video names are placed under each bar.}
    \label{fig:challenges}
\end{figure}
\begin{figure}[!t]
    \centering
    \centerline{\includegraphics[width=1\linewidth]{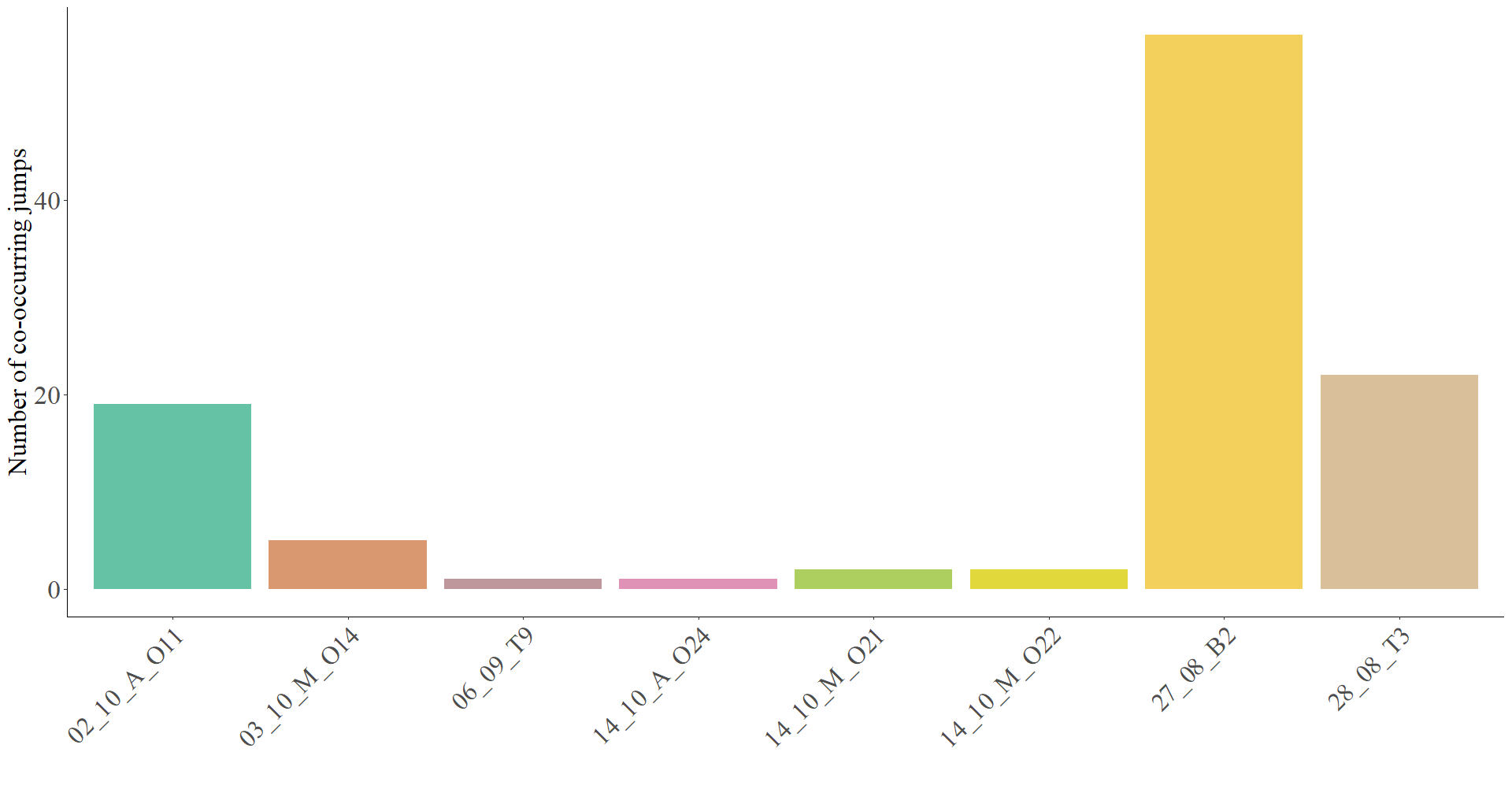}}
    \caption{Number of co-occurring jump events for each video. Original video names are placed under each bar.}
    \label{fig:cooccurring_jumps}
\end{figure}

\subsection{Tracking Challenges}

Tracking \textit{Trichogramma} wasps presents several unique challenges that distinguish WaspMOT from conventional MOT datasets.

First, individuals frequently perform abrupt jump events, resulting in large instantaneous position changes and temporary disappearance from detections. These events create trajectory fragmentation when trackers rely on motion continuity assumptions. The distribution of jump events is shown in Fig.~\ref{fig:challenges}, including cases where multiple individuals jump simultaneously (Fig.~\ref{fig:cooccurring_jumps}), increasing association ambiguity.

Second, occlusions occur due to interactions between individuals and the experimental arena structure. Because wasps can move both on the arena floor and ceiling, complete visual occlusions are possible even with a static camera setup. These occlusions can lead to temporary identity loss and incorrect reassignment. Fig.~\ref{fig:challenges} illustrates the distribution of occlusion events per sequence.

Third, individuals exhibit highly similar visual appearance, limiting the effectiveness of appearance-based discrimination. Combined with their small size and rapid motion, this makes reliable identity association particularly challenging.

Finally, the long duration of sequences requires trackers to maintain identity consistency over thousands of frames. Unlike conventional datasets where trajectories span only a fraction of the sequence, WaspMOT requires full-length trajectory reconstruction for accurate tracking.

These characteristics make WaspMOT a challenging benchmark for evaluating long-term tracking performance and identity preservation.

WaspMOT is not intended to replace large-scale MOT datasets with broader scene diversity. Instead, it complements them by targeting a distinct and underexplored evaluation regime: long-duration, closed-set tracking with full-length trajectories. In this regime, the primary challenge is not scene diversity or detector robustness, but maintaining identity consistency over thousands of frames.

\section{Benchmark Protocol}

\subsection{Oracle Detection Setup}

To isolate identity association performance from detection errors, we use oracle detections derived directly from ground truth bounding boxes without identity labels. This setup ensures perfect detection accuracy and allows evaluation to focus exclusively on the tracking component. All tracking results are generated in the MOTChallenge~\cite{mot17_ref} format and evaluated using the TrackEval~\cite{trackeval_ref} toolkit.

This design follows prior practice in scenarios where association is the primary factor of interest and enables a controlled evaluation of long-term identity preservation.

\subsection{Evaluated Tracking Algorithms}

We benchmark five tracking-by-detection methods representing different association strategies.

\textbf{ByteTrack}~\cite{bt_ref} associates detections using motion prediction based on a Kalman filter~\cite{kf_ref} and spatial overlap measured by intersection-over-union (IoU).

\textbf{BoT-SORT}~\cite{botsort_ref} extends this approach by incorporating appearance features alongside motion and spatial cues. Appearance embeddings are trained following the procedure described in the original work, adapted to WaspMOT.

\textbf{C-BIoU}~\cite{cbiou_ref} introduces an enhanced IoU-based matching strategy with enlarged bounding boxes, designed to improve robustness in environments with fast-moving subjects.

\textbf{OC-SORT}~\cite{ocsort_ref} improves motion modeling by emphasizing observation-centric updates, which enhances robustness under non-linear motion.

\textbf{McByte}~\cite{MCBYTE_REF} incorporates temporally propagated segmentation masks~\cite{cutie_ref,sam_ref} to provide additional geometric constraints during association.

All methods are applied using oracle detections and their default parameters without additional tuning. This setup enables consistent comparison across algorithms and focuses the evaluation on their inherent association capabilities.

\subsection{Spatial Tracklet Stitching Baseline}

To analyze recoverable identity fragmentation, we include a simple spatial tracklet stitching baseline applied as post-processing.

Each tracklet is represented by its endpoints in space and time. Candidate associations are formed by linking the end of one tracklet to the beginning of another based on spatial proximity and temporal consistency. When multiple candidates exist, global assignment is performed using the Hungarian algorithm~\cite{hungarianalg_ref} under one-to-one matching constraints.

This baseline does not introduce new modeling components but serves to quantify how much performance can be recovered from fragmented trajectories using simple spatial cues. It provides a reference point for evaluating the potential of more advanced long-term tracking methods.

\section{Results and Analysis}

\subsection{Evaluation Metrics}

We report three standard MOT metrics: HOTA~\cite{hota_ref}, IDF1~\cite{idf1_ref}, and MOTA~\cite{mota_ref}. HOTA captures overall tracking performance by combining detection and association quality, IDF1 measures identity consistency, and MOTA reflects detection accuracy and tracking errors. Higher values indicate better performance.

\subsection{Quantitative Results}

Table~\ref{tab:tracking_results} presents tracking performance of all evaluated methods, with and without spatial tracklet stitching.

All trackers exhibit substantial identity fragmentation, reflected in relatively low IDF1 scores despite the use of oracle detections. This confirms that long-term identity preservation remains challenging even when detection errors are eliminated.

Applying spatial stitching consistently improves performance across all methods. For example, ByteTrack improves by +3.8 HOTA and +8.8 IDF1, while McByte improves by +2.7 HOTA and +7.2 IDF1. Similar gains are observed for BoT-SORT, C-BIoU, and OC-SORT. These results indicate that a significant portion of identity fragmentation is recoverable using simple spatial consistency.

\begin{table}[!t]
\caption{Tracking performance of evaluated methods, before and after spatial tracklet stitching. Improvements over baseline are shown in parentheses.}
\centering
\small
\begin{tabular}{|c|c|c|c|}
\hline
Method & HOTA$\uparrow$ & IDF1$\uparrow$ & MOTA$\uparrow$ \\
\hline
ByteTrack~\cite{bt_ref} & 49.3 & 47.2 & 74.4 \\
ByteTrack + stitch & 53.1 (+3.8) & 56.0 (+8.8) & 74.4 (+0.0) \\
\hline
BoT-SORT~\cite{botsort_ref} & 47.9 & 45.4 & 74.4 \\
BoT-SORT + stitch & 51.7 (+3.8) & 54.4 (+9.0) & 74.5 (+0.1) \\
\hline
C-BIoU~\cite{cbiou_ref} & 51.3 & 50.5 & 74.4 \\
C-BIoU + stitch & 53.8 (+2.5) & 57.1 (+6.6) & 74.5 (+0.1) \\
\hline
OC-SORT~\cite{ocsort_ref} & 50.5 & 46.8 & 74.5 \\
OC-SORT + stitch & 54.5 (+4.0) & 55.5 (+8.7) & 74.5 (+0.0) \\
\hline
McByte~\cite{MCBYTE_REF} & 60.6 & 51.1 & 99.7 \\
McByte + stitch & \textbf{63.3} (+2.7) & \textbf{58.3} (+7.2) & \textbf{99.8} (+0.1) \\
\hline
\end{tabular}
\label{tab:tracking_results}
\end{table}

\begin{figure}[!t]
\centering
\begin{tabular}{ccc}
\includegraphics[height=3.2cm]{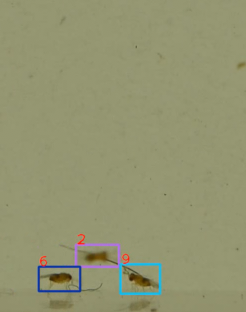}&
\includegraphics[height=3.2cm]{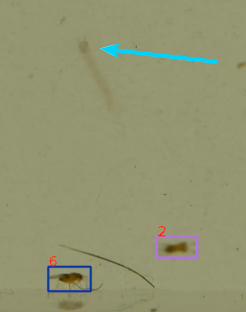}&
\includegraphics[height=3.2cm]{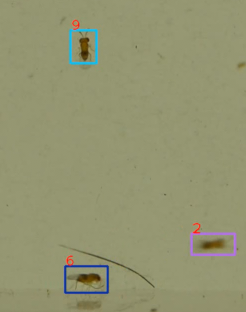}
\\
Frame 2232 & Frame 2238 & Frame 2241 \\
\end{tabular}
\caption{
An example of an individual \textit{Trichogramma} wasp performing an abrupt jump. The temporary disappearance results in trajectory fragmentation, which is recovered by spatial tracklet stitching.
}
\label{fig:wasps_jumping}
\end{figure}

MOTA remains largely stable across configurations due to the use of oracle detections, indicating that performance differences are driven primarily by identity association rather than detection quality. Notably, McByte achieves near-perfect MOTA, which can be attributed to its use of temporally propagated segmentation masks that help resolve ambiguous cases such as overlaps. In contrast, IDF1 shows the largest improvements for all algorithms, confirming that long-term identity consistency benefits most from trajectory reconstruction.

Among all methods, McByte~\cite{MCBYTE_REF} achieves the strongest overall performance, indicating that additional spatial constraints provided by mask propagation are beneficial. However, even the best-performing method fails to fully resolve identity fragmentation, highlighting the difficulty of long-term tracking.

The relative ordering of methods further reveals that appearance-based cues do not provide a clear advantage in this setting: BoT-SORT does not outperform simpler motion-based methods such as ByteTrack, despite incorporating learned embeddings. This suggests that appearance features are difficult to exploit effectively for small, visually similar targets. Overall, these results emphasize the complementary roles of motion, geometry, and long-term reasoning in WaspMOT.

\subsection{Qualitative Analysis}

Fig.~\ref{fig:wasps_jumping} illustrates a typical failure case caused by an abrupt jump event, where an individual temporarily disappears and its trajectory is fragmented. Spatial stitching successfully reconnects the trajectory, demonstrating that such errors are not due to ambiguity in identity but to limitations of local association mechanisms.


Overall, these results demonstrate that current tracking-by-detection methods struggle to maintain identity consistency over long durations. The consistent improvements obtained with simple spatial stitching suggest that substantial gains remain possible, motivating the development of methods explicitly designed for long-term identity preservation.

\section{Conclusion}

We introduced WaspMOT, a multi-object tracking benchmark designed to evaluate long-term identity preservation in ecological surveillance scenarios. The dataset consists of fully annotated long-duration sequences with oracle detections, enabling controlled evaluation of identity association over full-length trajectories.

Through a benchmark of five tracking-by-detection methods, we showed that current approaches struggle to maintain consistent identities over extended durations, even in the absence of detection errors. Consistent improvements obtained with a simple spatial stitching baseline indicate that a significant portion of trajectory fragmentation is recoverable, but also highlight the limitations of existing association strategies.

WaspMOT exposes a failure mode of modern trackers that is not captured by conventional short-duration benchmarks and provides a foundation for future research on long-term identity preservation. The dataset and code will be publicly released upon acceptance to support further development in long-term tracking and ecological monitoring applications.

\section*{Acknowledgement}
This work has been supported by the French government, through the 3IA Cote d’Azur Investments in the project managed by the National Research Agency (ANR) with the reference number ANR-23-IACL-0001. 

This work was granted access to the HPC resources of IDRIS under the allocation 2025-AD011014370 made by GENCI.

\bibliographystyle{IEEEtran}
\bibliography{references}

\end{document}